%% file: survey.tex
\documentclass[11pt,a4paper]{article}
\usepackage[utf8]{inputenc}
\usepackage{amsmath}
\usepackage{amsfonts}
\usepackage{amssymb}
\usepackage{graphicx}

\usepackage[ruled]{algorithm2e}
\renewcommand{\algorithmcfname}{ALGORITHM}
\SetAlFnt{\small}
\SetAlCapFnt{\small}
\SetAlCapNameFnt{\small}
\SetAlCapHSkip{0pt}
\IncMargin{-\parindent}

\usepackage{fullpage}
\usepackage{color}
\usepackage{authblk}

\usepackage{url}
\usepackage{epsfig}
\usepackage{bm}
\usepackage{amsmath}
\usepackage{multirow}
\usepackage{subfigure}
\newcommand{\eat}[1]{}

\begin{document}

\title{A Survey on Social Media Anomaly Detection}
\date{}
\author[1]{Rose Yu%
	\thanks{Electronic address: \texttt{qiyu@usc.edu}; Corresponding author}}
\author[1]{Huida Qiu}
\affil[1]{USC}
\author[2]{Zhen Wen}
\affil[2]{Google}
\author[3]{Ching-Yung Lin}
\affil[3]{IBM Research}
\author[1]{Yan Liu}
\setcounter{Maxaffil}{0}
\renewcommand\Affilfont{\itshape\small}
\maketitle

\begin{abstract}
Social media anomaly detection is of critical importance to prevent malicious activities such as bullying, terrorist attack planning, and fraud information dissemination. With the recent popularity of social media, new types of anomalous behaviors arise, causing concerns from various parties. While a large amount of work have been dedicated to traditional anomaly detection problems, we observe a surge of research interests in the new realm of social media anomaly detection.  In this paper, we present a survey on existing approaches to address this problem. We focus on the new type of anomalous phenomena in the social media and review the recent developed techniques to detect those special types of anomalies. We provide a general overview of the problem domain, common formulations, existing methodologies and potential directions.  With this work, we hope to call out the attention from the research community on this challenging problem and open up new directions that we can contribute in the future.
\end{abstract}

%

%
%

\section{Introduction}
\input{intro.tex}

\section{Point Anomaly Detection}
Point anomaly refers to the abnormal behaviors of individual users, which can be reflected in abnormal activity records such as unusually frequent access to important system files, or abnormal network communication  patterns. Point anomaly detection aims to detect suspicious individuals, whose behavioral patterns deviate significantly from the general public. Based on the type of input, we can have activity-based point anomaly detection and graph-based point anomaly detection.

\subsection{Activity-based Point Anomaly Detection}

\input{individual_act.tex}

\subsection{Graph-based Point Anomaly Detection}
\input{individual_graph.tex}

\section{Group Anomaly Detection}
Group anomaly or \textit{``collective anomaly''} detection in social network aims to discover groups of participants that collectively behave anomalously \cite {chandola2007outlier}. This is a challenging task due to three reasons: (1) we do not know beforehand any members of a malicious group; (2) the members of anomalous groups may change over time; (3) usually no anomaly can be detected when we examine individual member. Most existing algorithms can only address one or two of these challenges.

\subsection{Activity-based Group Anomaly Detection}
\input{group_act.tex}

\subsection{Graph-based Group Anomaly Detection}
\input{group_graph.tex}

\section{Conclusions and Future Research}
\input{conclusions.tex}

\section{Acknowledgment}
\input{acknowledge.tex}


%

\bibliographystyle{plain}
\bibliography{ref}
\end{document}

%% file: intro.tex
Social media systems provide convenient platforms for people to share, communicate, and collaborate. While people enjoy the openness and convenience of social media, many malicious behaviors, such as bullying, terrorist attack planning, and fraud information dissemination, can happen. Therefore, it is extremely important that we can detect these abnormal activities as \textit{accurately} and \textit{early} as possible to prevent disasters and attacks. 
Needless to say, as more social information becomes available, the most challenging question is what useful patterns could be extracted from this influx of social media data to help with the detection task. 

By definition, anomaly detection aims to find ``an observation that deviates so much from other observations as to arouse suspicion that it was generated by a different mechanism'' \cite{Hawkins1980}. 
The common approach is to build a reference model, i.e., a statistical model that captures the generation process of the observed (or normal) data. Then for a new observation, we estimate its likelihood based on the reference model and predict the data as an ``anomal'' if the likelihood is below some threshold \cite{Chan_stolfo_1998,Ghosh_schwartzbard_1999,Eskin02ageometric,Ringberg_2007,Yue_Wu_Wang_Li_Chu_2007,Chandola:2007kx}.

In addition, the type of anomalies that we aim to detect vary significantly from applications to applications. 
Several algorithms have been developed specifically for social network anomaly detection on graph structure anomalies, e.g. power law models \cite{Akoglu2009}, spectral decomposition \cite{Luxburg2007}, scan statistics \cite{priebe05scan}, random walks \cite{Pan:2004:AMC:1014052.1014135,Tong:2006:FRW:1193207.1193363}, etc. The basic assumption of these algorithms is that if a social network has fundamentally changed in some important way, it is usually reflected in the individual communication change, i.e., some individuals either communicate more (or less) frequently than usual, or communicate with unusual individuals. 
However, this could be an over-simplification of the social media anomalies without considering several important aspects of social media data.

One of the challenges that differentiate social media analysis from existing tasks in general text and graph mining is the \textit{social layer} associated with the data. In other words, the texts are attached to individual users, recording his/her opinions or activities. The networks also have social semantics, with its formation governed by the fundamental laws of social behaviors. The other special aspect of social media data is the \textit{temporal perspective}. That is, the texts are usually time-sensitive and the networks evolve over time. Both challenges raise open research problems in machine learning and data mining. Most existing work on social media anomaly detection have been focused on the social perspective. For example, many algorithms have been developed to reveal hubs/authorities, centrality, and communities from graphs \cite{Kleinberg99,Erosheva04b,Song05,Lappas09}; a good body of text mining techniques are examined to reveal insights from user-generated contents \cite{Blei03,Rosen-Zvi04}. However, very few models are available to capture the temporal aspects of the problem \cite{blei06,Hanneke2010,Kolar10}, and among them even fewer are practical for large-scale applications due to the more complex nature of time series data.
 
Existing work on traditional anomaly detection \cite{chandola_TKDD_2010,ChengTPK09,Chandola:2007kx,TongPSYF08,Guralnik_1999,Takeuchi_TKDD_2006,Duch_CORR_2004,Lin_DMKD_2003,Keogh2005HotSax,Yankov2008diskAware,ChengTPK09,WSARE3} have identified two types of anomalies: one is ``univariate anomaly'' which refers to the anomaly that occurs only within individual variable, the other is ``dependency anomaly'' that occurs due to the changes of temporal dependencies between time series. Mapping to social media analysis scenario, we can recognize two major types of anomalies: 
\begin{itemize}
\item Point Anomaly: the abnormal behaviors of individual users
\item Group Anomaly: the unusual patterns of groups of people
\end{itemize}

Examples of point anomaly can be anomalous computer users \cite{schonlau2001computer}, unusual online meetings \cite{horn2011online} or suspicious traffic events \cite{Ihler:2006AED}. Most of the existing work have been devoted to detecting point anomaly. However, in social network, anomalies may not only appear as individual users, but also as a group. For instance, a set of users collude to create false product reviews or threat campaign in social media platforms; in large organizations malfunctioning teams or even insider groups closely coordinate with each other to achieve a malicious goal. Group anomaly is usually more subtle than individual anomaly. At the individual level, the activities might appear to be normal \cite{chandola2007outlier}. Therefore, existing anomaly detection algorithms usually fail when the anomaly is related to a group rather than individuals. 

We categorize a broad range of work on social media anomaly detection with respect three criteria: 
\begin{enumerate}
\item Anomaly Type: whether the paper detects point anomaly or group anomaly
\item Input Format: whether the paper deals with activity data or graph data
\item Temporal Factor: whether the paper handles the dynamics of the social network
\end{enumerate} 

In the remaining of this paper, we organize the existing literature according to these three criteria. The overall structure of our survey paper is listed in table \ref{tb:structure}. We acknowledge that the papers we analyze in this survey are only a few examples in the rich literature of social media anomaly detection. References within the paragraphs and the cited papers provide broader lists of the related work.
\begin{table}[ht]
 \label{tb:structure}
\caption{Survey Structure }
\begin{tabular}{c  c}
\hline
\multicolumn{2}{c}{\multirow{2}{*}{\textbf{Point Anomaly Detection}} } \\
& \\
\hline
\multirow{2}{*}{Activity-based}& \textit{Bayes one-step Markov}, \textit{compression}\cite{schonlau2001computer},  \textit{multi-step Markov} \cite{Vardi99ahybrid}, \\    
&    \textit{Poisson process} \cite{Ihler:2006AED},   \textit{probabilistic suffix tree} \cite{sun2006mining} \\
\hline
\multirow{2}{*}{Graph-based } &  \textit{random walk} \cite{outrank08,neighborFormationADinBGraphICDM05},  \\ 
&   \textit{power law} \cite{Akoglu2009,akoglu2010oddball} \\ 
\multirow{2}{*}{(static graph) } &\textit{hypergraph} \cite{silva2008hypergraph,Silva:2008iss}\\
&  \textit{spatial autocorrelation}\cite{sun2004local,chawla2006slom} \\
\hline
Graph-based  & \textit{scan statistics} \cite{priebe05scan,park2008scan}, \textit{ARMA process}     \cite{Lakhina2004NetTrafAD} \\
(dynamic graph) &   \textit{MDL}   \cite{graphScopeKDD07,akoglu2012fast}, \textit{graph eigenvector} \cite{eigenSpaceADKDD04} \\
\hline
\multicolumn{2}{c}{\multirow{2}{*}{\textbf{Group Anomaly Detection}} } \\
& \\
\hline
\multirow{2}{*}{Activity-based}&  \textit{scan statistics} \cite{das2009detecting,findingTribesKDD07}, \textit{causal approach} \cite{babbar2013causal}  \\
& \textit{density estimation} \cite{xiong2011hierarchical,xiong_group_2011,muandet2013one,yu2014glad} \\
\hline
\multirow{2}{*}{Graph-based} & \textit{MDL} \cite{autopartPKDD04,linkDiscoveryViaRarityICDM03,anomalousLinkKDDExplor05} \\
& \textit{anomalous substructure} \cite{noble2003graph,anomStuctSimToNormICDM07} \\
(static graph)  & \textit{tensor decomposition} \cite{maruhashi2011multiaspectforensics} \\
\hline
Graph-based& \textit{random walk} \cite{changingSubgraphsICDM08}, \textit{t-partitie graph} \cite{SCANKDD07,CHRONICLE09}\\
(dynamic graph) & \textit{counting process} \cite{Heard:2010zr} \\
\hline
\end{tabular}
\end{table}%

We can also formulate the categorization in Table \ref{tb:structure} using the following mathematical abstraction. Denote the time-dependent social network as $G=\{V(t), W_v(t), E(t), W_e(t)\}$, where $V$ is the graph vertex, $W_v$ is the weight on the vertex, $E$ is the graph edge and $W_e$ is the weight on the edge. Point anomaly detection learns an outlier function mapping from the graph to certain sufficient statistics $F:G \rightarrow R$. A node is anomalous if it lies in the tail of the sufficient statistics distribution. Group anomaly detection learns an outlier function mapping from the power set of the graph to certain sufficient statistics $F:2^{|G|} \rightarrow R$. Activity based anomaly detection collapses the edge set $E(t)$ and weights $W_e(t)$ to be empty. Static graph-based approaches fix the time stamp of the graphs as one. Now each of the method summarized in the table is essentially learning a different $F$ or using some projection (simplification)
of the graph $G$. The projection trades-off between model complexity and learning efficiency.

%% file: individual_act.tex
User activities are widely observed in social media, such as computer log-in/log-off records, HTTP access records, and file access records. Activity-based approaches assume that individuals are marginally independent from each other. The anomalousness of an individual is determined only by his own activities. A large body of literature are in the context of computer intrusion detection study. For example,  \cite {schonlau2001computer}  investigates the problem of detecting masquerades who disguise themselves as somebody else 
on the network. The paper collects user activities by looking at their UNIX commands records and manipulating the data to simulate masquerades. 

Pioneering work for detecting masquerades fall into the framework of statistical hypothesis testing, e.g. \cite{Dumouchel99Bayes1-stepMarkov,Vardi99ahybrid}. Different approaches are proposed including \textbf{Bayes one-step Markov}, \textbf{hybrid multi-step Markov} and \textbf{compression}. Here we omit other simple masquerade detection techniques such as uniqueness of the command as also compared in \cite{Dumouchel99Bayes1-stepMarkov}.  For Bayes one-step Markov method, it states the null hypothesis as a one-step Markov process and the alternative hypothesis as a Dirichlet distribution. The null hypothesis assumes that the current time command $C_{ut}$ of a user $u$  relates to his previous command $C_{u,t-1}$. Mathematically speaking, $H_0:P(C_{ut}=k|C_{u,t-1}=j) =  p_{ukj}$, where $p_{ukj}$ is the transition probability from command $j$ to command $k$ for user $u$. Then the algorithm computes the Bayes factor based on the hypothesis for each user $\bar{x}_u$ and set up a threshold with respect to $\bar{x}_u$ to detect anomalous masquerades. This approach models users independently and ignores the potential relationships among users.

As a direct generalization of Bayes one-step Markov,  \cite{Vardi99ahybrid} builds a user model based on high-order Markov chains: \textbf{hybrid multi-step Markov}. It tests over two hypotheses. $H_0:$ commands are generated from the hybrid Markov model of $u$; $H_1:$ commands are generated from other users. The hybrid multi-step Markov method switches between the Markov model and the independence model. The Markov model assumes that a command depends on a set of previous commands, i.e. $P(C_{ut}=c_0 | C_{u,t-1} =c_1,C_{u,t-2}=c_2,\cdots,C_{u,t-l}=c_l )= \sum_{i=1}^l \lambda_{ui}r_u(c_0|c_i)$, where $\lambda$ and $r$ denotes the initial and transitional probability.  For the independence model, it assumes that a user's commands are i.i.d samples from a multinomial distribution. The paper computes the test statistics by combining the statistics from two models. Similar to Bayes one-step Markov, hybrid multi-step Markov method sets up a threshold value on the test statistics to flag anomalies. Hybrid multi-step Markov method is able to capture the long-range dependence of the users' commands. However, it also suffers from higher computational cost.\textbf{ Compression} takes a distinctive approach where it defines the anomaly score as the additional compression cost to append the test data to the training data. Formally, the score is $x = \text{compress}(\{C, c\}) - \text{compress}(C)$, where $C$ is the training data, $c$ is the testing data. The method applies the Lempel-Ziv algorithm for the compression operation. However, it can hardly capture the dependencies in the data instances.

\begin{figure}[hbtp]
\centering
\includegraphics[scale=1]{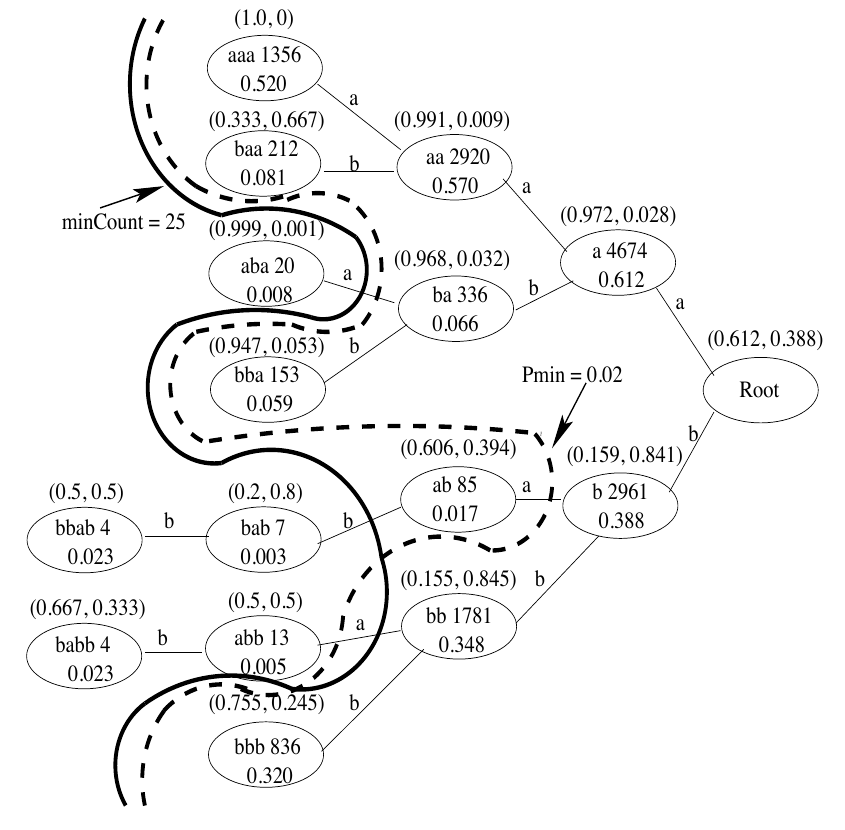}
\caption{A example of PST. For each node, top array shows the probability distribution. Inside the node shows the label string, the number of times it appears in the data set and the empirical probability. \cite{sun2006mining}}\label{fig:PST}
\end{figure}

\cite{sun2006mining} proposes \textbf{probabilistic suffix tree} (PST) to mine the outliers in a set of sequences $S$ from an alphabet $\Sigma$. It makes Markov assumption on the sequences and encodes the variable length Markov chains with syntax similar to Probabilistic Suffix Automata. In PST, an edge is a symbol in the alphabet and a node is labeled by a string. The probabilistic distribution of each node represents the conditional probability of seeing a symbol right after the string label. An example of such PST is shown in Figure \ref{fig:PST}. The algorithm  first constructs a PST and then computes a similarity measure score $SIM_N$ based on marginal  probability of each sequence over the PST. Then it selects the top $k$ sequences with lowest $SIM_N$ scores as outliers. Since PST encodes a Markov chain, which has been shown to have certain equivalence to the Hidden Markov model, the outliers detected by PST are similar to those using Markov model testing statistics. Though PST construction and $SIM_N$ are relative cheap in computation, one drawback is that PST is pre-computed for a fixed alphabet. Pre-computation makes PST less adaptive to the unseen symbols outside of the alphabet or newly coming sequences, which basically requires recomputing the entire tree.

\vspace{-0.2in}

\cite{Ihler:2006AED} investigates Markov-modulated \textbf{Poisson process} to address the specific problem of event detection on time-series of count data. The algorithm assumes the count at time $t$, denoted as $N(t)$, is a sum of two additive processes: $N(t) = N_0(t) +  N_E(t)$, where $N_0(t)$ denotes the number of occurrences attributed to the ``normal'' behavior and $N_E(t)$ is the  ``anomalous'' count due to an  event at time $t$. More concretely, the periodic portion of the time series counts can be taken as ``normal'' behavior while the rare increase in the number of counts can correspond to the ``anomalous''  behavior.  For both processes, the paper develops a hierarchical Bayesian model. In particular, the paper models periodic counting data (i.e. normal behavior) with a Poisson process and models rare occurrences (i.e. anomaly behavior) via a binary process. The algorithm then uses the MCMC sampling  algorithm to infer the posterior marginal distribution over events.  It uses the posterior probability as an indicator to automatically detect the presence of unusual events in the observation sequence. The paper applies the model to detect the events  from free-way traffic counts and the building access count data. The method takes a full Bayesian approach as a principled way to pose hypothesis testing. However, it treats each time series as independent and fails  to consider the scenario where multiple time series are inter-correlated.

Another application in social network anomaly detection is proposed in \cite{horn2011online}. The paper proposes to detect unusual meetings by investigating the presence of meeting participants. Specifically, for each time stamp $t=1,2,\cdots$, the inputs are given as a snapshot of the network in the form of a binary string $x_t = (x_t(1), \cdots, x_t(n) ) \in \{0,1\}^{n}$, where $x_t(j) =0$ or $1$ indicates whether the $j$th person participated in the meeting at time $t$ as well as the feedback from expert system with the correct labels $y_t \in \{-1,+1\}$. The algorithm outputs a binary label for each network state $\hat{y}_t \in \{-1,+1\}$ according to whether or not $x_t$ is anomalous.   Under the proposed two-stage framework, ``filtering stage'' estimates the model parameters and updates belief with the new observation. It builds an exponentially model driven by a time-vary parameter and learns the model parameter in an online fashion.  ``Hedging stage'' compares the model likelihood of $x_t$ as   $\zeta_t$ with the critical threshold  $\tau_t$ and flag anomalies if $\zeta_t > \tau_t$. After that, the online learning algorithm utilizes the feedback from an expert system to adjust the critical threshold value $\tau_{t+1} =  \text{argmin}_{\tau} (\tau -  \tau_t - \eta y_t 1_{\hat{y_t} \neq y_t})^2$. It is easy to see from the construction of $x_t$ that each person's participation is taken as an independent feature entry. Though this work highlights the network structure, the relational information utilized  lies  only between people and meetings, without considering the interaction among people themselves.

Generally speaking, activity-based approaches model the activity sequence of each user separately under certain Markov assumption. They locate the anomaly by flagging deviations from a user's past history. These approaches provide simple and effective ways to model user activities in a real-time fashion. The models leverage the tool of Bayesian hypothesis testing and detect anomalies that are statistically well-justified. However, as non-parametric methods, Markov models suffer from the rapid explosion in the dimension of the parameter space.  The estimation of Markov transition probabilities becomes non-trivial for large scale data set. Furthermore, models for individual normal/abnormal activities are often ad hoc and are hard to generalize. As summarized by  \cite{Schonlau2000uniqueness} in his review work on computer intrusion detection: ``none of the methods described here could sensibly serve as the sole means of detecting computer intrusion''. Therefore, exploration of deeper underlying structure of the data with fast learning algorithms is necessary to the development of the problem. Here we also refer interested readers to more general reviews of computer network anomaly detection \cite{Ahmed:2007vn,Lazarevic:2003ys}.

%% file: individual_graph.tex
Social media contain a considerably large amount of relational information such as emails from-and-to communication, tweet/re-tweet actions and mention-in-tweet networks. Those relational information are usually represented by graphs.  Some approaches analyze static graphs, each of which is essentially one snapshot of the social network.  Others go beyond static graph and analyze dynamic graphs, which is a series of snapshot of the networks.

\subsubsection{Static Graph}

Compared with activity-based approaches, which simplify the social network as categorical or sequential activities of individuals, graph-based approaches further take into account the relational information represented by the graph. \cite{noble2003graph} immerses as one of the earliest work focusing on graph-based anomaly detection. It introduces two techniques for graph-based anomaly detection. One is to detect anomalous substructures within a graph and the other is to detect unusual patterns in distinct sets of vertices (subgraphs). Substructure is a connected component in the overall graph. Subgraph is obtained by partitioning the graph into distinct structures. Each substructure is evaluated using the Minimum Description Length metric for anomalousness.  In real social graphs, intensive research efforts have been devoted to study the graph properties (see references in \cite{graphLawGenAlgo06}). One famous example is the power law, which describes the relationship among various attributes, namely the number of nodes ($N$), number of edges ($E$), total weight ($W$) and the largest eigen-value of the adjacency matrix ($\lambda$).
\eat{
\begin{definition}[Power Law]
\cite{graphLawGenAlgo06} Two variables $x$ and $y$ are related by a power law when:
$$y(x)=Ax^{-\gamma}$$
where $A$ and $\gamma$ are positive constants. The constant $\gamma$ is often called the power law exponent.
\end{definition}}

Based on these observations, \cite{Akoglu2009} proposes to study each node by looking at \textbf{power law}  property in the domain of its ``egonet'', which is the subgraph of the node and its direct neighbors. For a given graph $\mathcal{G}$, denote the egonet of node $i$  as $\mathcal{G}_i$, the paper describes the ``OddBall'' algorithm.  The algorithm starts by investigating the number of nodes  $N_i$, the weight $W_i$ and number of edges $E_i$ of the egonet $\mathcal{G}_i$. It then defines the normal neighborhoods patterns with respect to these quantities. For example, the authors report the Egonet Density Power Law (EDPL)  pattern for $N_i$ and $E_i$: $E_i \propto N_i^\alpha, \quad 1 \leq \alpha \leq 2.$ ;  the Egonet Weight Power Law (EWPL) pattern for $W_i$ and $E_i^\beta, \quad \beta \geq 1$. Given the normal patterns, the paper takes the distance-to-fitting-line as a measure to score the nodes in the graph.  The algorithm can detect anomalous nodes whose neighbors are either too sparse (Near-star) or too dense (Near-clique).  By studying both the total weight $W$ and the number of edges $E$, it can detect anomalous nodes whose interactions with others are extremely intensive. By analyzing the relationship between the largest eigenvalue $\lambda$ and the total weight $W$, it can detect dominant heavy link, or a single highly active link in an egonet. The ``OddBall'' algorithm builds on  power law properties of complex networks, which haven been verified in various real world applications. Moreover, the fitting of power law and the calculation of anomaly score is computationally efficient, which makes the algorithm a good fit for large scale network analysis. However, the algorithm would easily fail if the network does not obey the power law, then the detected anomalies would be less meaningful. Also, the paper focuses only on the static network and generalization the algorithm to dynamic network is non-trivial.

Besides the power law, \textbf{random walk} is also adapted for graph-based anomaly detection between neighbors. The general idea is that if a node is hard to reach during the random walk, it is likely to be an anomaly. Random walk calculates a steady state probability vector, each element of which represents the probability of reaching other nodes. Following the idea of random walk, \cite{neighborFormationADinBGraphICDM05} focuses on the anomaly detection on bipartite graph, denoted as $G= \langle V_1 \bigcup V_2, E \rangle$, where node sets $V_1$ has $k$ nodes, $V_2$ has $n$ nodes and $E$ are the edges between them. It detects anomalies by first forming the neighborhood and then computing the normality scores. During neighborhood formation stage, the algorithm computes the relevance score for a node $b\in V_1$ to $a \in V_1$ as the number of times that one visit $b$ during multiple random walks starting from $a$. In this case, the steady state vector represents the probability of being reached from $V_1$ in a \textit{random walk with restart} model, and the algorithm detects anomalies linked to the query nodes. Random walk model stresses the graph structure while ignores the nodes' attributes. Sometimes, it might be an over-simplification of the underlying network generating process, which would lead to high false positive ratio. 

 \cite{outrank08} uses similar random walk guideline to detect outliers  in a database and  proposes the 	``OutRank'' algorithm. It first constructs a graph from the objects where each node represents a data object and each edge represents the  similarity between them. For  every pair of the objects $X, Y \in \mathbb{R}^d$, the algorithm computes the similarity $Sim(X, Y)$ and normalizes the resulting similarity matrix to obtain a random walk transition matrix $S$.  Then it defines the following connectivity metric  based on how well this node is connected to the other nodes: 

\textit{Definition.(Connectivity)} Connectivity $c(u)$ of node $u$ at $t$th iteration is defined as follows:

\begin{equation*}
c_{t}(u) =
\begin{cases}
   a &\mbox{if $t=0$ }\\
   \sum\limits_{v \in \text{adj}(u)}(c_{t-1}(v) /|v|) & \text{otherwise}
   \end{cases}
\end{equation*}

where $a$ is its initial value, $\text{adj}(u)$ is the set of nodes linked to node $u$, and $|v|$ is the node degree. This recursive definition of connectivity is also known as the power method for solving eigenvector problem. Upon convergence, the stationary distribution can be written as $c = S^Tc$.  The algorithm detects the objects (nodes) with low connectivity to other objects as anomalies. ``OutRank'' solves individual activity-based anomaly detection problem using  a graph-based anomaly detection method. As a general outlier detection framework, it requires the construction of the graph from data objects. Thus its performance can heavily rely on the type of similarity measurement adopted for computing the edges.

\begin{figure}[hbtp]
\centering
\includegraphics[scale=1]{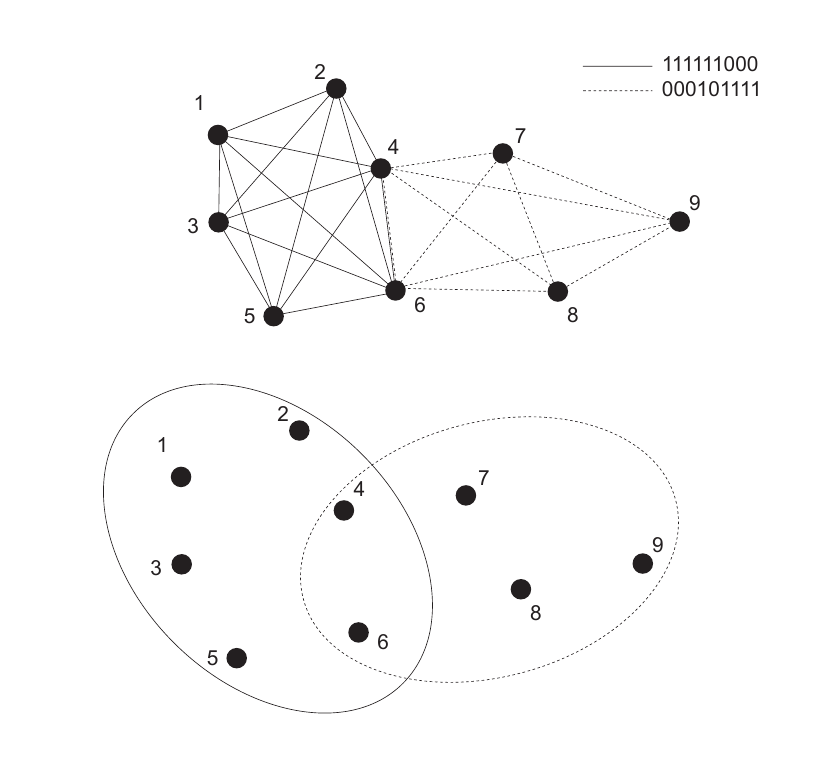}
\caption{Modeling two observations, 111111000 and 000101111, with p = 9, using a graph (top) and a hypergraph (bottom). With the graph, representing one observation of an interaction requires multiple edges. With a hypergraph, one hyperedge suffices. The hypergraph is more efficient for storing/representing observations and more informative about the real structure of the data. \cite{silva2008hypergraph}}\label{fig:hypergraph}
\end{figure}

Despite a wealth of theoretical work in graph theory, standard graph representation only allows each edge to connect to two nodes, which cannot encode potentially critical information regarding how ensembles of networked nodes interacting with each other \cite{silva2008hypergraph}. Given this consideration, a generalized \textbf{hypergraph} representation is formulated  which allows edges to connect with multiple vertices simultaneously. In hypergraph, each hyperedge is a representation of a binary string,  indicating whether the corresponding vertex participates in the hyperedge.  Figure \ref{fig:hypergraph} provides an example for comparing the graph and the hypergraph representation of two observations 111111000 and 000101111, with p = 9, using a graph and a hypergraph. With the graph, representing one observation of an interaction requires multiple edges. With a hypergraph, one hyperedge suffices. Due to the mapping between binary strings and hyperedges, the paper formulates the graph-based anomaly detection  problem in the corresponding discrete space.  \cite{silva2008hypergraph} and \cite{Silva:2008iss} address the problem of detecting anomalous meetings in very large social networks based on hypergraphs. In their papers, a  meeting is encoded as a hyperedge $\mathbf{x}$ and $g(\mathbf{x}) $ is the probability mass function of the meetings evaluated at $\mathbf{x}$.  The distribution of the meetings is modeled as a two-component mixture of a non-anomalous distribution and an anomalous event distribution $g(\mathbf{x}) = (1-\pi) f(\mathbf{x}) + \pi \mu(\mathbf{x})$, with $\pi$ as the mixture parameter. Then the  paper learns the likelihood of each observation using variational EM algorithm with a multivariate Bernoulli variational approximation.  The likelihood is subsequently used for the  evaluation of the anomalousness. Hypergraph is specifically designed for high dimensional data  in the graph. It provides a concise representation of the complex interactions among multiple nodes. But the representation only applies to binary relationships where an edge is either present or missing.

\cite{sun2004local,chawla2006slom} consider using  \textbf{spatial auto-correlation}  to detect local spatial outliers. We categorize them as graph-based approach because the spatial neighborhood defined in those methods resembles the neighborhood defined in graph. For each point $o$, the paper defines the Spatial Local Outlier Measure (SLOM) as $\tilde{d}(o) * \beta(o)$ to score its anomalousness. According to their definition, $\tilde{d}$ is the ``stretched'' distance  between the point and its neighbors and $\beta$ is the oscillating parameter. SLOM captures the spatial autocorrelation using $\tilde{d}$ and spatial heteroscedasticity(non-constant variance) with $\beta$. However, when the data is of high dimensions, the concept of neighborhood becomes less well-defined. The local anomaly defined in the proposed method using local spatial statistics would suffer from the ``curse of dimensionality''.

Generally speaking, static graph-based approaches consider not only the activity of individual users but also their interactions. The common practice is to extract important node features from the graph, which relies heavily on feature engineering. Some algorithms import graph theoretical properties such as the power-law or the random walk into the analysis. However, those approaches usually make strong assumptions on the graph generating process, which can be easily violated in real world social networks.

\subsubsection {Dynamic Graph}
Social networks are dynamic in nature. Therefore, it is worthwhile to consider the problem of anomaly detection in a dynamic setting. A brief survey on dynamic network anomaly detection is elaborated in \cite{bilgin2010dynamic}. The survey characterizes the techniques employed for the problem into three groups: \textit{Time Series Analysis of Graph Data, Anomaly Detection using Minimum Description Length, Window Based Approaches}. Based on this categorization, we review those anomaly detection approaches that incorporate the network dynamics into their models.

Dynamic networks can be represented as a time series of graphs. A common practice is to  construct a time series from the graph observations or substructures. \cite{ADinSeriesGraphsARMA05} uses a number of
graph topology distance measures to quantify the differences between two consecutive networks, such as weight, edge, vertex, and diameter. For each of these graph topology distance measures,  a time series of changes
is constructed by comparing the graph for a given period with the graph(s) from one or
more previous periods. Given a graph $G= \{V,E, W_V, W_E\}$, the algorithm constructs a  time series of changes for each graph topology distance measures. Each time series is individually modeled by an \textbf{ARMA process}. The anomaly is defined as days with
residuals of more than two standard errors
from the best ARMA model. The paper detects anomalies by setting up a residual threshold for the goodness of model fitting for time series.  The proposed method in \cite{ADinSeriesGraphsARMA05} is designed for change point detection. The performance of the proposed algorithm  highly depends on how the graph topology distance measures are defined. Additionally, the distance measure is  only able to capture the correlation between two consecutive time stamps rather than long-range dependencies.

\textbf{Graph eigenvectors} of the adjacency matrices is another form of  the time series extracted from dynamic graph streams. In \cite{eigenSpaceADKDD04}, the paper addresses the problem of anomaly detection in computer systems. Assume a system has $N$ services,  the paper defines a time evolving dependency matrix $D \in \mathbb{R}^{N\times N}$, where each element of the matrix $D_{i, j}$ is a function value relate to the number of service $i$'s requests for service $j$ within a pre-determined time interval. Given a time series of dependency matrices $D(t)$, the algorithm extracts the principal eigenvector $\mathbf{u}(t)$ of $D(t)$ as the ``activity'' vector, which can be interpreted as the distribution of the probability that a service is holding the control token of the system at a virtual time point. To detect anomalies, the authors define the typical pattern as a linear combination of the past activity vectors  $\mathbf{r}(t) =  c\sum_{i=1}W v_i \mathbf{u}(t-i+1)$, where $\{v_i\}$ are the coefficients and $c$ is the normalization constant.  Then the algorithm calculates the dissimilarity of the present activity vector from this typical pattern. The anomaly metric $z(t)$ is defined as $z(t) =1- \mathbf{r}(t-1)^T \mathbf{u}(t)$. When the anomaly metric quantity $z(t)$ is greater than a given threshold, the system flags anomalous situation.  Compared with representing graphs with edges, weights and vertices as in \cite{ADinSeriesGraphsARMA05}, features built upon eigenvectors capture the underlying invariant characteristics of the system and preserve good properties such as positivity, non-degeneracy, etc.

\eat{According to their arguments, the feature vector is robust in presence of constant factor changes in the adjacency matrix (change from $M$ to $k\cdot M$ where $k$ is a nonzero real number), and represents the stationary state of a discrete-time dynamical system defined by the adjacency matrix. Thus, from the graph series we get a series of feature vectors $\{\bm{\mu} (t)\}$, which forms a series of matrix $U(t)$:
$$U(t)=[\bm{\mu}(t), \bm{\mu}(t-1), ..., \bm{\mu}(t-W+1)]$$
Where $W$ is a window size. The typical pattern of the graphs in a window is defined as a linear combination:

\begin{align}
& \bm{r}(t)  =  c\sum_{i=1}^{W}{v_i \bm{\mu}(t-i+1)}=cU(t)\bm{v}(t) \\
& \mbox{ and } \bm{v}(t) = \arg\max_{\bm{v}}{\|\sum_{i=1}^{W}v_i\bm{\mu}(t-i+1)\|^2} \\
&              \mbox{ subject to } \bm{v}^T\bm{v}=1 \nonumber
\end{align}
where $c$ is the normalization constant to satisfy $\bm{r}^T\bm{r}=1$. The following metric is defined to evaluate the dissimilarity between two consecutive windows:
$$z(t)=1-\bm{r}(t-1)^T\bm{\mu}(t)$$
Which is assumed to follow the von Mises-Fisher distribution. A significance level is manually chosen to get a threshold $z_{th}$ and alert is flagged when $z(t)>z_{th}$.}

\begin{figure}[hbtp]
\centering
\includegraphics[scale=1]{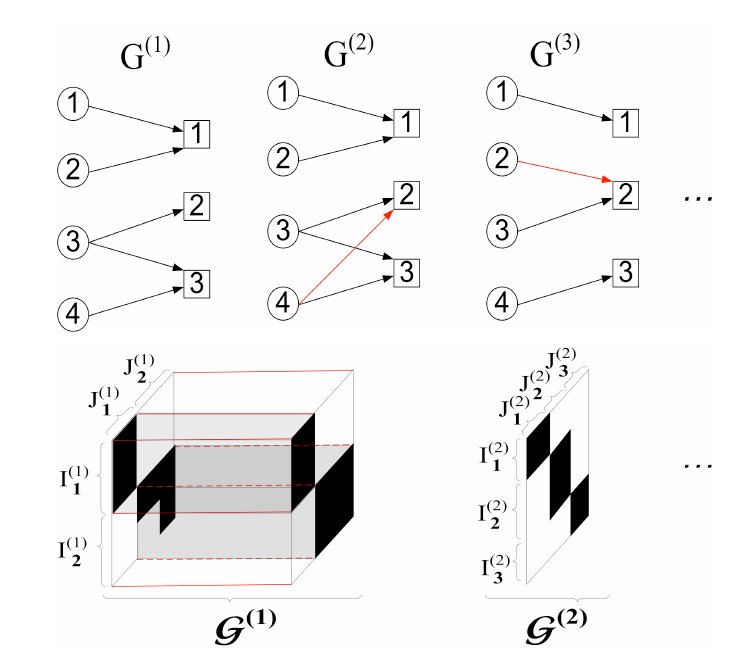}
\caption{A graph stream with 3 graphs in 2 segments. First graph segment consisting of 
	$G^{(1)}$ and $G^{(2)}$   
	has two source partitions $ I^{(1)}_1 = \{1, 2\}$, $I^{(1)}_2= \{3,4\}$; two
	destination partitions $J^{(1)}_1 = \{1\}$, $J^{(1)} = _2 = \{2, 3\}$. Second graph segment consisting of $G^{(3)}$ has three source partitions $I^{(2)}_1 = \{1\}$, $I^{(2)}_2 = \{2, 3\}$, $I^{(2)}_3 = \{4\}$; three destination partitions $J^{(2)}_1 = \{1\}$, $J^{(2)}_2 = \{2\}$, $J^{(2)}_3 = \{3\}$. \cite{graphScopeKDD07}}\label{fig:segment}
\end{figure}

Besides time series analysis of the  graph stream, \textbf{Minimum description length} (MDL) has been applied to anomaly detection as another way of characterizing the dynamic networks. \cite{graphScopeKDD07} detects the change points in a stream of graph series. It introduces the concept of graph segment, which is one or more graph snapshots and the concept of source/destination partitions, which groups the source and destination nodes into clusters. Figure \ref{fig:segment} illustrates those concepts in a three graph series.  The rational behind the algorithm is to consider whether it is easier to include a new graph into the current graph segment or to start a new graph segment. If a new graph segment is created, it is treated as a change point. Given current graph segment $\mathcal{G}^{(s)}$, encoding cost $c_o$ and a new graph $G^{(t)}$, the algorithm computes the encoding cost for $\mathcal{G}^{(s)}\bigcup \{G^{(t)}\}$ as  $c_n$ and $G^{(t)}$ as  $c$. If $c_n-c_o<c$, the new graph is included in the current segment. Otherwise, $\{G^{(t)}\}$ forms a new stream segment and time $t$ is a change point. To compute the encoding cost of a graph segment, the algorithm tries to partition the nodes in a segment into source and destination nodes so that the MDL for encoding the partitions is minimized. In this case, a change point indicates the time when the structure of the graph has dramatically changed. One limitation of this algorithm is that it can only handle unweighted graphs, which cannot encode the intensity of the communication between users. Thus, this method does not fit the situation when the communications of people suddenly increase while the topological structure stays unchanged. (e.g. a heated discussion starting to prevail in a social network).

\vspace{-0.2in}
\vspace{-0.1in}

\cite{akoglu2012fast} addresses the categorical anomaly detection by pattern-based compression, which also adopts MDL-principle. It encodes a database with multiple code tables and searches for the best partitioning of features using MDL-optimal rule. With the natural property of code tables, the algorithm declares the anomaly by the pattern that has long code word, which are rarely used and have high compression cost. The method has been successfully generalized to a broad range of data. The use of multiple code tables to describe the data in the proposed algorithm exploits the correlations between groups of features. But the partition of the features into groups would impose unrealistic independence assumptions on the data.

For window-based approach, \textbf{scan statistics} is the main-stream  method.  The idea of scan statistics is to slide a small window over local regions, computing 
certain local statistic (number of events for a point pattern, or average
pixel value for an image) for each window. The supremum or maximum of these
locality statistics is known as the scan statistic. \cite{priebe05scan} specifically discusses a framework of using  scan statistics to perform anomaly detection on dynamic graphs. Specifically, the algorithm defines the scan region by considering the closed $k$th-order neighborhood of vertex $v$ in graph $D =(V,E)$: $N_k[v;D]=\{w\in V(D): d(v,w) \leq k\}$. Here distance $d(v,w)$ is the minimum directed path length from $v$ to $w$ in $D$. The induced subdigraph $\Omega(N)(N_k[v;D])$ is thus the scan region and any digraph invariant $\Psi_k(v)$ of the scan region is the locality statistics.  For instance, the out degree of the digraph can be one such invariant locality statistics.  The scan statistic $M_k(D)$ is the maximum locality statistic over all vertices. The algorithm applies hypothesis testing by stating the null hypothesis (normality) and the alternative hypothesis (anomaly). Digraphs with large scan statistic indicates the existence of the anomalous activity and are rejected under null hypothesis with certain threshold.  Extension of scan statistics from standard graph to hypergraph representation is also examined in \cite{park2008scan} for time-evolving graphs. The scan statistic is an intuitively appealing method to evaluate dynamic graph patterns. But one drawback of this type of method is the necessity to pre-specify a window width before one looks at the data.

%% file: group_act.tex
Activity-based group anomaly detection approaches usually assume that the group information is given beforehand and devote the most effort to model the activities within groups. Those approaches also imply that groups are marginal independent with each other.

\cite{das2008anomaly} proposes a probabilistic model to detect group of anomalies in categorical data sets. It generalizes the spatial \textbf{scan statistic} in \cite{priebe05scan} for dynamic graphs to non-spatial data sets  with discrete valued attributes. It uses Bayesian networks to model the relationship between the attributes and computes the group score for all subsets of the data $S$ based on the model likelihood: $F(S)= \frac{P(Data|H_1(S)}{P(Data|H_0)}$. Under this definition, $H_0$ is the null hypothesis that no anomalies are present, and $H_1(S)$ is the alternative hypothesis specifying subset $S$ is an anomalous group. Then it performs a heuristic search over arbitrary subsets of the data to find the groups that maximize the likelihood. At the final stage, it performs randomization testing to evaluate the statistical significance of the detected groups. For spatial data, the computation of scan statistics involves a definition of scanning region, which is often based on geographical properties. Non-spatial categorical data has the difficulty in defining local statistics based on geographical properties. Therefore, the efficient search heuristic is critical to the performance of algorithm. On the other hand, it lacks the solid theoretical justification and is sensitive to model mis-specification. 

\cite{das2009detecting} considers the anomalies in categorical data sets and tries to detect anomalous attributes or combinations of attributes. The paper proposes two algorithms to test for anomalous records, i.e \textit{Conditional Probability Test} and \textit{Marginal Probability Test}. Conditional probability test uses conditional probability as the testing statistic. For two attributes $a_t, b_t$, the algorithm considers the ration $r(a_t, b_t) =\frac{P(a_t)P(b_t)}{P(a_t, b_t)}$. Marginal probability computes a quantity called the q-value, which is the cumulative probability mass of all the attributes $\text{q-val}(a_t) = \sum_{x\in X} P(x)$ where $X = \{x:P(x) \leq P(a_t)\}$. Q-value is in parallel with the p-value. This approach concerns with empirical distribution functions and is parameter free. But the underlying distributions of the attributes would heavily depend on the sample size of the data.

\begin{figure}[htbp]
\centering
\subfigure[Multinomial Genre Model]{\includegraphics[scale =1.2]{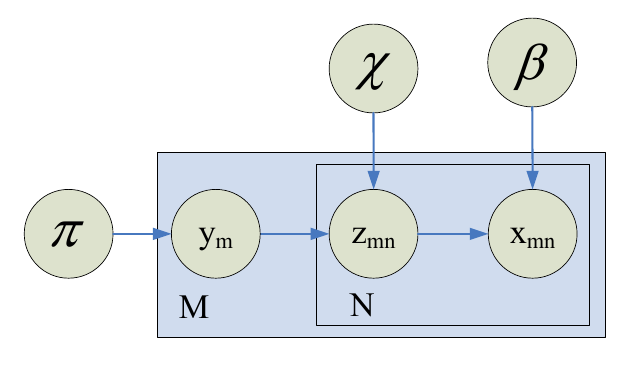}}\label{fig:MGM}
\subfigure[Flexible Genre Model]{\includegraphics[scale =1.4]{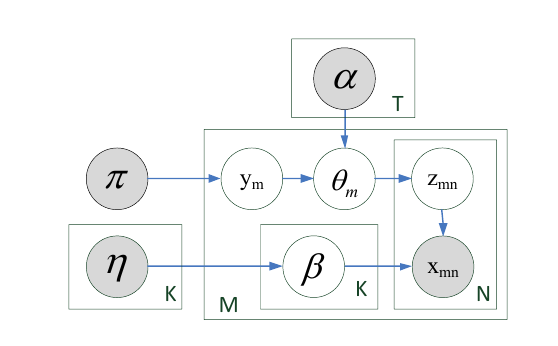}}
\caption{Graphical Model Representation of Multinomial Genre Model and Flexible Genre Model for activity-based group anomaly detection.}\label{fig:FGM}
\end{figure}

Another line of work formulates the group anomaly detection problem as a \textbf{density estimation} task. It imposes a hierarchical probabilistic model on the normal groups and estimates the distribution of the latent variables in the model. It evaluates the likelihood of the estimated latent variables for individual group and use it as a test statistic. The Multinomial Genre Model (MGM) proposed in  \cite{xiong2011hierarchical} first investigates the problem following the paradigm of latent models.  MGM models groups as a mixture of Gaussian distributions with different mixture rates. Formally, given $M$ groups, each of which has $N_m$ objects.  MGM assumes that the object features $X_{m.n}$ are generated from a mixture of $K$ Gaussian, $m = 1,2,\cdots, M$, $n = 1,2,\cdots, N_m$ with a set of stereotypical mixture rates $\chi$.  The mixture rates of the $M$ groups belong to one of the stereotypical mixture rates in $\chi$.  Figure \ref{fig:MGM} depicts the graphical model of the proposed model. The method then performs Bayesian inference to estimate the density of the mixture rate for each group. Then group anomaly detection is conducted by scoring the mixture rate likelihood of each group.  This method finds groups whose topic variables $\{Z_m, n\}$ are not compatible with any of the stereo- typical topic distributions in $\chi$. In MGM, groups share the candidate topics $\beta$ globally, which leads to bad performance when groups have different sets of topics. \cite{xiong_group_2011} further extends MGM to Flexible Genre Model (FGM)  with more flexibility in the generation of topic distributions, as shown in Figure \ref{fig:FGM}. The motivation of FGM is to allow  each group to have its own topics.  Specifically, they change the set of topics $\beta$ from model hyper-parameters to random variables, depending on the genre parameter $\eta$. This extension enables the model  to adapt to more diverse genres in groups.

Apart from the generative approach used in MGM and FGM,  \cite{muandet2013one} takes a discriminative approach to estimate the density of the mixture model.  It uses the same definition of group anomaly from \cite{xiong2011hierarchical} and represents groups as probability distributions. The authors consider kernel embedding of those probabilistic distributions. For two probabilities $\mathbb{P}_1$ and $\mathbb{P}_2$, the kernel on probability distributions is defined as $K(\mathbb{P}_1, \mathbb{P}_2)= \int\int k(x, y) d\mathbb{P}_i(x) d\mathbb{P}_j(y)$, where $k$ is a reproducing kernel in reproducing kernel Hilbert space (RKHS). They generalize the technique of one-class support vector machine (OCSVM) for point anomaly detection to group anomaly detection. Similar to OCSVM with translation invariant kernels, the authors compute the kernel of Gaussian distributions and apply SVM in a probability measure space. Interestingly, the proposed one class support measure machine (OCSMM) algorithm has inherent correspondence to the kernel density estimation, which is theoretically more attractive. Compared with generative approaches in \cite{xiong2011hierarchical,xiong_group_2011}, OCSMM does not make assumptions on the underlying distribution of the data and is generally less computational expensive. However, due to the use of Gaussian RBF kernels and support vector machine, the algorithm is inevitably sensitive to the selection of kernels as well as the soft margin parameter.
\vspace{-0.2in}
\vspace{-0.1in}
\cite{babbar2013causal} takes a casual approach to detect the contextual anomaly. The paper proposes to encode the variables in the Bayesian network and use probabilistic association rule to discover anomalies. The association rule builds upon two measures namely \textit{support} and \textit{confidence}.  Support describes the prior probability of a variable while confidence represents the conditional probability.  Given a state variable $X$ and observations $Y$, the paper defines the two measures as follows $suppport(X = x_i) = P(X=x_i)$ and $confidence(X=x_i) = Pa(X=x_i|Y)$, where $Pa$ is the parent nodes of $X$ in the Bayesian network.  The algorithm detects the domain specific anomalous patterns (DSAP) based on two probabilistic association rules: 1) low support and high confidence 2) high support and low confidence. Then it sorts the detected DSAPs according to sensitivity analysis  scores and considers the top $\tau$ patterns with the lowest scores as output anomalies. Different from MGM or FGM, the proposed method operates on the general Bayesian network rather than a specific probabilistic model. The evaluation of support and confidence on each node is relatively cheap compared with full Bayesian inference. However, the detected causal anomalies would be ad hoc. The false positive rate would increase sharply with larger size  Bayesian networks.

\eat{Group anomaly detection problem can also be found in multivariate time series observations. For the activities in time series, \cite{huida2012granger} uses a regularized \textbf{Granger causality} graphical model to capture the characteristics of time series. After learning the causal graph by regularization, they compute anomaly score by calculating the KL divergence of reference set distribution and the test set distribution. The group anomalies are determined by threshold cutoff the anomaly scores.}

%% file: group_graph.tex
The most common observations we have in social networks are the individual attributes as well as ties among participants. Graph-based group anomaly detection techniques seek to jointly utilize these observations and detect anomalous groups in a unified framework.

\subsubsection{Static Graph}

Anomalous edge detection has been proposed in \cite{autopartPKDD04} based on graph partitioning. The algorithm aims to detect anomalous edges that deviate from the overall clustering structure. The rationale behind this method is that if the removal of an edge  can significantly make the graph easier to partition, then the two linked nodes may have an anomalous relation. The partitioning algorithm tries to find the best number of partitions so that the \textbf{Minimal Description Length} (MDL) needed to encode and transmit all the partitions of the graph is minimized. For a graph with $n$ nodes, the paper defines the group mapping $\mathcal{G}:\{1,2,..., n\} \rightarrow \{1,2,\cdots, k\}$ to assign nodes into $k$ clusters. Thus, the \textit{Total Encoding Cost}  for the graph $T(D;k,\mathcal{G})$ in the form of  the adjacency matrix $D=[d_{i, j}]$ depends on the number of the clusters $k$ as well as the group mapping of the nodes $\mathcal{G}$. Anomalous edges are those edges whose removal would significantly reduce the total graph encoding cost. In the paper, the anomaly score of an edge is defined as the total encoding cost  difference to transmit the new partitions when the edge is removed, i.e, ``outlierness'' of edge $(u, v):= T(D';k,\mathcal{G}) - T(D;k,\mathcal{G})$. $D$ and $D'$ are equal of all edges except that $d'_{u, v}=0$. Other similar work includes \cite{linkDiscoveryViaRarityICDM03}, which defines a \textit{rarity} measure to discover unusual links, and \cite{anomalousLinkKDDExplor05}, which uses a \textit{Katz} measurement to statistically predict the likelihood of a link.  Edge anomaly detection focus merely on pair-wise relationship and is not feasible for detecting more complicated anomalous behaviors with more than two people involved.

Finding \textbf{anomalous substructure} in graphs is another topic of attention.  For example, in the scenario of email exchanges within a company, email correspondence between managers and their secretaries should be normal (frequent pattern), while email exchange between assembly line workers and secretaries could be an anomalous pattern. \cite{noble2003graph} presents an iterative expanding algorithm to look for rare substructures using their SubDue system \cite{SubDue2000}. Given a labeled graph, where each node has a label identifying its type, the system starts with a list holding $1$-vertex substructures for each unique vertex label. It modifies the list by generating, extending, deleting or inserting vertices and edges. One central issue is how to measure the anomalousness of a substructure. Simply counting the number of occurrences for substructures is not enough, as larger substructures tend to have low occurrences. \cite{noble2003graph} intuitively defines a score for a substructure $S$ in a graph $G$ as $F_2 = Size(S)\cdot Occurrences(S, G)$, which is simply the product of the total number of nodes within a substructure and its occurrences. A smaller value of $F_2$ indicates a more abnormal substructure. Another issue of the problem is the computational complexity of the algorithm. Although \cite{SubDue2000} shows that in practice the system runs in polynomial time, theoretically it faces exponential number of substructures.

The pioneering work of \cite{noble2003graph} sees the rise of mining substructures in graphs.  \cite{maruhashi2011multiaspectforensics} leverages the structural information in the heterogeneous networks to detect unusual subgraph patterns. The algorithm encodes the graph using a tensor and focuses on finding the suspicious spikes via \textbf{tensor decomposition}.
Formally, given an M-mode tensor $\mathcal{X}$ of size $I_1 \times I_2 \times \cdots \times I_M$, the algorithm performs CP decomposition of the tensor of rank $R$ as $\mathcal{X} \approx \sum_{r=1}^R \lambda_r (a_r^{(1)} \times \cdots  a_r^{(M)})$, where $\{a_r^{(i)}\}$ are rank-1 eigenscore vectors. The approximation would be exact when $R$ equals the true rank of the tensor. Next the algorithm transforms the eigenscore vector plot (absolute value of eigenscore vs. attribute index) into the eigenscore histogram (absolute value of eigenscore vs. frequency count) and conducts spike detection on the histogram. The proposed approaches bridges graph mining and tensor analysis. Tensor decomposition is able to capture the complex structure in heterogeneous networks. But tensor decomposition problem itself can be NP-hard to solve. And the lack of explicit objective in the proposed anomaly detection framework would create difficulties in the final evaluation of the algorithm's performance.
 
In the setting of fraudulent activity detection, \cite{anomStuctSimToNormICDM07} jointly considers anomalous substructure and the criteria of MDL. Specifically, they run the SUBDUE system with MDL heuristics to find the normative pattern in the graph. Instances of substructure are evaluated against the normative pattern with a match cost. Anomalous substructures are the ones with the lowest matches. Based on this definition of group anomaly, \cite{anomStuctSimToNormICDM07} presents three slightly different algorithms, i.e. GBAD-MDL, GBAD-P and GBAD-MPS to detect anomalies. These methods first find all the instances of frequent substructures and evaluate the frequency of the abnormal structure multiplied by the match cost.  A key drawback of this method is that it assumes that the degree of nodes in a graph is uniformly distributed, which is almost impossible in most social networks.  As shown in \cite{weightedGraphPatternKDD08,graphLawGenAlgo06}, real graphs usually follow power law degree distribution instead of uniform distribution.

\begin{figure}[htbp]
\centering
\includegraphics[scale = 0.4]{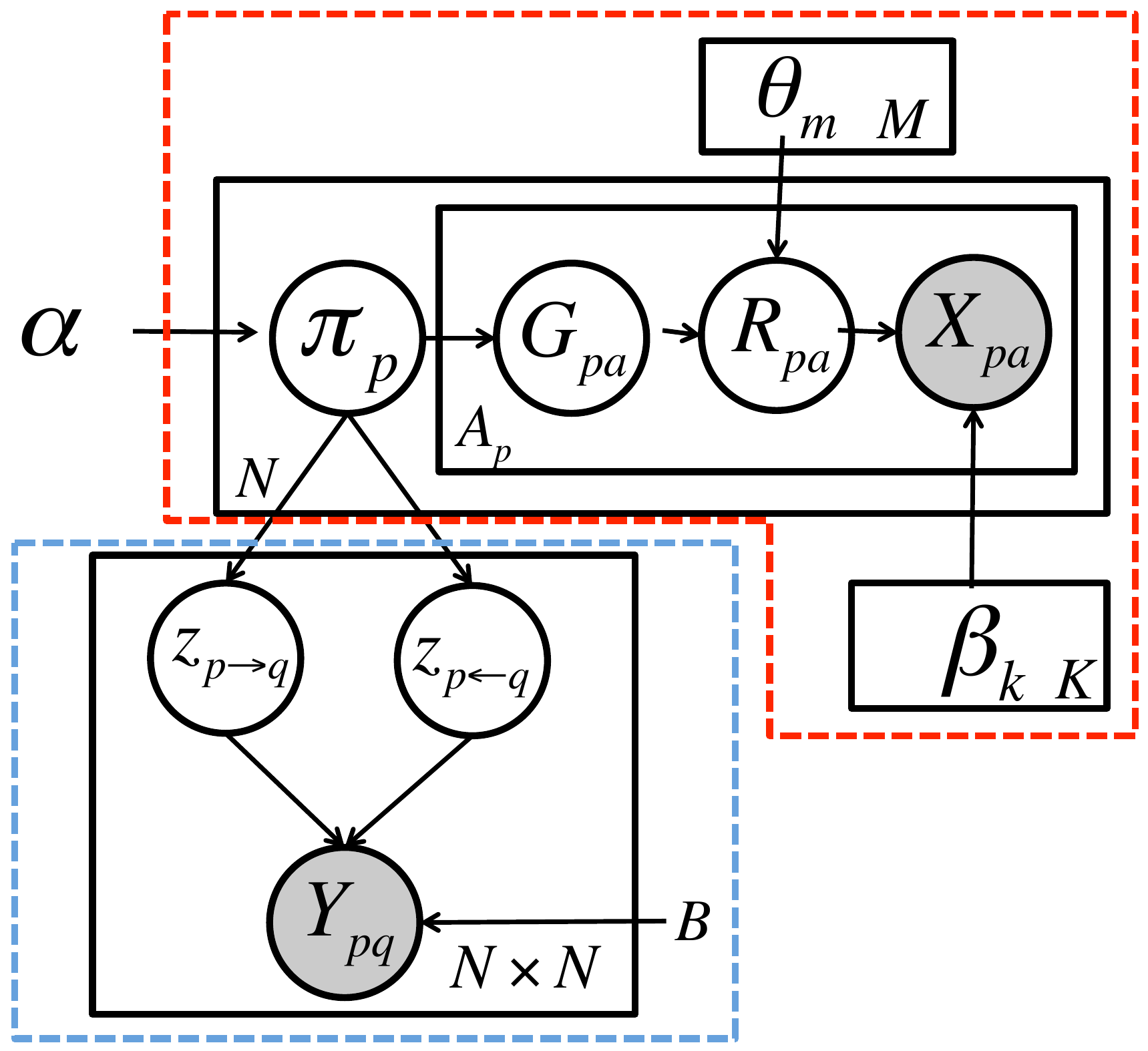}
\caption{Plate representation for the GLAD model}
\label{fig:graphGLAD}
\end{figure}

In social media, two forms of data coexist: one is the point-wise data, which characterize the features of an individual person. The other is pair-wise relational data, which describe the properties of social ties. Density estimation methods for group anomaly detection \cite{xiong2011hierarchical,xiong_group_2011,muandet2013one} emphasize on the point-wise   data and usually overlook the pair-wise relational data.  Graph-based methods highlight the graph structure but usually fail to account for the attributes of individual nodes. Additionally, existing group anomaly detections algorithms are all two-stage approaches: (\romannumeral 1) identify groups, (\romannumeral 2) detect group anomalies. This strategy assumes that the point-wise and pair-wise data are marginally independent. However, such independence assumption might underestimate the mutual influence between the group structure and the  feature attributes. The detected group anomalies can hardly reveal the joint effect of these two forms of data. 

With those considerations, \cite{yu2014glad} proposes to build an alla prima that can accomplish the tasks of group discovery and anomaly detection all at once. They develop a hierarchical Bayes model: the  GLAD model, for detecting the group anomaly. The GLAD model utilizes both the pair-wise and point-wise data and automatically infers the group membership and the role at the same time. It models a social network with $N$ individuals. Assuming that each person $p$ is associated with a group identity $G_p$ and a role identity $R_p$. By groups, it means the clusters that capture the similarity suggested by the pair-wise communications. By roles, it refers to the mixture components that categorize the point-wise feature values of the nodes. For simplification, they fix the number of groups as $M$ and the number of roles as $K$. Figure \ref{fig:graphGLAD} shows the plate notation for the GLAD model.

For each person $p$, he joins a group according to the membership probability distribution $\pi_p$. GLAD imposes a Dirichlet prior on the membership distribution. It is well known that the Dirichlet distribution is conjugate to the multinomial distribution. It assumes the pair-wise link $Y_{p,q}$ between person $p$ and person $q$ depends on the group identities of both $p$ and $q$ with the parameter $B$. Furthermore, it models the dependency between the group and the role using a multinomial distribution parameterized by a set of role mixture rate $\{\theta_{1:M}\}$. The role mixture rate characterizes the constitution of the group: the proportion of the population that plays the same role in the group. Finally, it models the activity feature vector of the individual $X_p$ as the dependent variable of his role with parameter set $\{\beta_{1:K}\}$. 

GLAD defines the group anomaly based on the role mixture rates, it scores the group anomalousness using $-\sum_{p \in G}\langle\log p(R_p|\Theta)\rangle_p$. The most anomalous group will have the highest anomaly score. In practice, it approximates the true log likelihood with the variational log likelihood to get $-\sum_{p \in G} \langle\log p(R_p|\Theta)\rangle_q$. A limitation of GLAD is that it only models the static network. This might be restrictive if we want to further consider dynamic networks. Besides the anomaly group whose mixture rate deviates significantly from other groups, it is also interesting to study how the mixture rate evolves over time.

\subsubsection{Dynamic Graph}
Evolving networks can also provide insights into the temporal changes of groups. Detecting anomalously groups in dynamic graphs is more challenging, as the group structures are not fixed and the unusual patterns in the group can also change. 

\cite{findingTribesKDD07} take a bipartite graph of individual entities and sequential ordered attributes as inputs and returns a group of entities whose attributes sequences are less likely to be generated from the proposed Markov chain model. One example of this type of anomaly is that several people constantly jump from companies to companies together. They track the complete history of employments and disclosures, and recognize the tribes that are closely related. Formally, the method requires bipartitie graph $G=(R\bigcup A, E)$, where $R =\{r_{i}\}$ is the entity representatives, $A=\{a_j\}$ is the attributes and $E$ are edges with time interval annotation. For each edge $e \in E$, $e=(r_i, a_j, tstart_{ij}, tend_{ij})$.  The method begins by listing the co-worker relationships in the graph. Every pair $f_{ij} = (r_i,r_j)$ indicates the individuals that have worked together. This results in a new graph $H=(R, F)$, where edges in the new graph $F=\{f_{ij}\}$ is annotated with individuals attribute and history information.  Then the paper defines a significance score for each edge, which measures the significance or the anomalousness of shared jobs.  The algorithm proceeds by identifying significant edges and computing the significance  score $c$ for each of them. Then the proposed method picks a threshold $d$ for the scores and prune all the edges $f_{ij}$ for $c_{ij} <d$. After pruning, the connected components in the remaining graph (which should be quite sparse after the pruning) are regarded as anomalous groups, or tribes as referred in the paper. As also pointed by authors, the choice of scoring pairs constitutes the heart of the problem, thus posing difficulty in the selection 

\cite{proximTrackEvolvBGraphSDM08} directly analyze graph structures  and  efficiently track node proximity, which measures the relevance between two nodes in bipartite graphs. The paper defines a dynamic proximity score based on the probability to ``random walk'' from one to the other in the static graph. Low proximity to other nodes can in a way indicate anomaly. Their definition of dynamic proximity accounts for two important aspects of node relevance: proximity involves multiple snapshots of the graph; proximity does not drop over time. \cite{proximTrackEvolvBGraphSDM08} extends this method to track anomalous nodes in time evolving graphs by defining a dynamic proximity metric. This dynamic proximity is derived from the edge and weight differences between graph snapshots and preserves a monotonicity property.

\cite{changingSubgraphsICDM08} proposes to detect the significant changing subgraphs. Given two consecutive snapshots of a graph $G_{i-1}$ and $G_i$, the algorithm defines an \textit{importance score} to measure the accumulative  change of a node's closeness to its $l$-step neighbors (neighbors within $l$ hops from the node) between two consecutive graph slices. In their context, \textbf{random walk} with restart is used to  model the node relevance. The closeness of a pair of vertices $v_j$ and $v_k$ is defined as $\Pi^l(j,k) = \sum_{\tau:v_j =\rightarrow v_k; \text{length}(\tau) \leq l} p(\tau) c(1-c) ^{\textit{length}(\tau)} $, where $\tau$ is a path from $v_j$ to $v_k$ whose length is $\text{length}(\tau)$ with transition probability $p(\tau)$. The importance score is therefore the summation of the closeness changes of $v_j$ to the other nodes, defined as $VI_i(v_j) = \sum_{v_k \in V_i} |\Pi_{i-1}^l (j,k) -  \Pi_i^l (j,k)|$ . Note that two consecutive graph slices $G_i$ and $G_{i+1}$ have the same set of nodes, but their edge set could be different. With the node closeness $\Pi^l_{i}$ and the vertex importance score $VI$, the paper uses a strategy similar to density clustering to detect the significant subgraphs. Specifically, the algorithm puts the most important node in the current subgraph $g$, adds all of its $l$-step neighbors to a max-heap. As long as there exists a node whose closeness with node $t$ exceeds certain threshold, the algorithm iteratively moves $t$ from the heap into $g$. When the iteration terminates, $g$ is regarded as the anomalous subgraph, and the algorithm proceeds to generate anomalous subgraphs for the next timestamp. The proposed algorithm detects subgraphs with significant change in edges as  group anomalies. The incremental learning of nodes closeness changes makes the algorithm quite efficiently. However, the output subgraphs heavily rely on the threshold for the closeness, and there is no clear mapping between the nodes' closeness and anomalousness.

\begin{figure}[ht]
\centering
    \includegraphics[scale= 1.2]{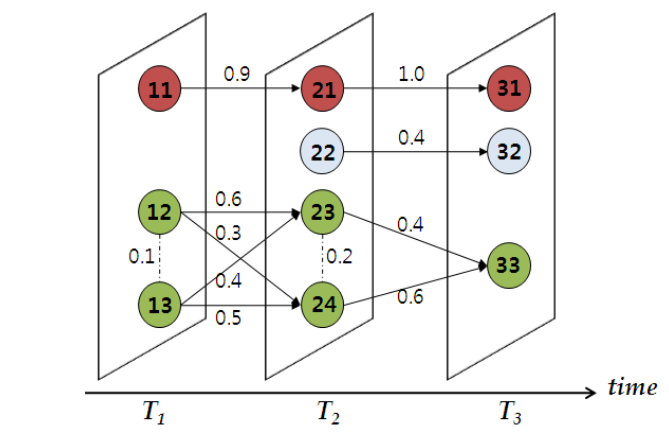}
\caption{An example of $t$-partite graph constructed from a dynamic network (taken from \cite{CHRONICLE09}). Each circle at a time stamp $T_i$ represents a cluster in the snapshot graph $G_i$.}
\label{figure:t_partite}
\end{figure}
\vspace{-0.2in}

\begin{figure}[htbp]
\centering
\includegraphics[scale =0.6 ]{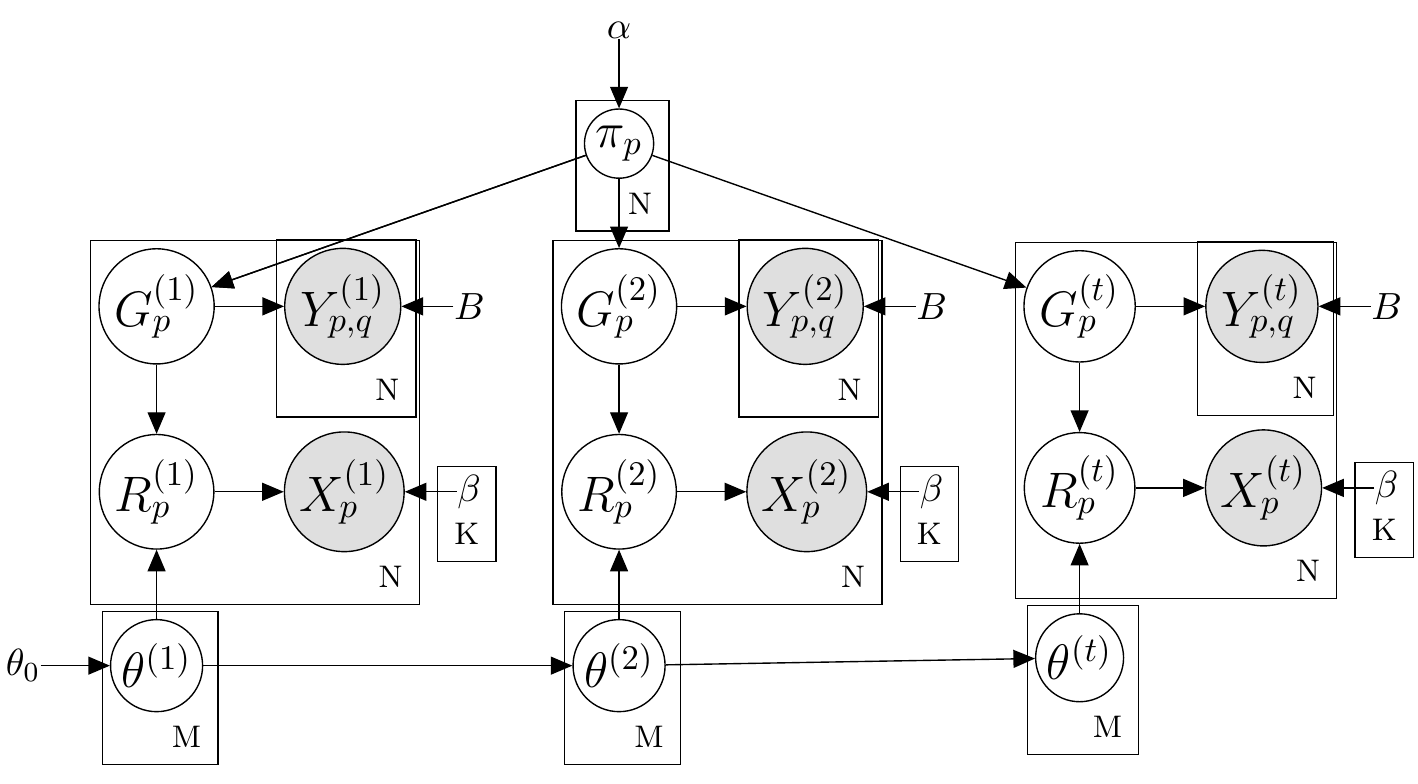}
\caption{Plate representation for the Dynamic Group Latent Anomaly Detection (DGLAD) model}
\label{fig:graphDGLAD}
\end{figure}
\vspace{-0.2in}
In another work which tries to detect changing communities by \cite{CHRONICLE09}, the authors propose a two-stage density-based clustering algorithm \textit{CHRONICLE}. The algorithm first clusters nodes in each snapshot graph $G_i$ at time $T_i$ using \textit{structural similarity} $\sigma$, defined as $\sigma(v,w)= \frac{|N(v) \bigcap N(w)|}{ \sqrt{|N(v)| \times |N(w)|}}$, where $N(v)$ is the neighborhood nodes of $v$. Then the algorithm replaces each cluster with one node to form a \textbf{$t$-partite graph}, as shown in Fig. \ref{figure:t_partite}. In the $t$-partite graph, the edge weight between two nodes (dashed edges in Fig. \ref{figure:t_partite}) within a time stamp $T_i$ denotes the number of edges between the two clusters, and the edge weight between two nodes from two consecutive time stamps is defined as the \textit{Jaccard} similarity between the node sets of two clusters. In the second stage, \textit{SCAN} is applied again on the $t$-partite graph. In Fig. \ref{figure:t_partite} different colors represents different clusters in the $t$-partite graph. From these clusters we can clearly monitor the formation and dissolving of the groups, which could provide some hint on which groups are anomalously changing. However, the algorithm is originally designed for monitoring community evolution instead of anomalous changing group detection. It is not clear how the algorithm can be adapted for anomaly detection yet.

\cite{Heard:2010zr} also presents a two-stage method, which combines the Bayesian approach for  discrete activity modeling with the graph analysis techniques for discovering anomalous structures. In the first stage identifies potentially anomalous nodes by conjugating Bayesian models for discrete time \textbf{counting processes}. Specifically, it models   the number of communications made from $i$ to $j$ up until discrete time $t$ denoted by $ N_{ij}(t)$ as a counting process.  It learns the distributions of the counts and use predictive p-value to evaluate the new observations for anomalous nodes. In the second stage, standard network inference tools are applied to the reduced subnetwork of the anomalous nodes identified from the first stage to uncover anomalous structure. Simulated cell phone communication as well as real-time press and media summary data are investigated to validate the method. This approach does not distinguish between point anomaly and group anomaly, hence hard to evaluate.

To further account for the dynamic nature of social media, \cite{yu2014glad} generalizes GLAD to the d-GLAD model as an extension for handling time series and formulate the problem as a change point detection task. The paper models the temporal evolution of the role mixture rate for each group with a series of multivariate Gaussian distributions. At a particular time point, the Gaussian has its mean as the value of the mixture rate. And the mixture rate of the next time point is a normalized sample from this Gaussian distribution. Since the model requires the mixture rate to be the parameters of a multivariate distribution over features, the authors apply a soft-max function to normalize the sample drawn from the multivariate Gaussian. The soft-max function is defined as $S(\theta_m) = \frac{\exp \theta_m}{\sum\limits_m \exp \theta_m}$. When the total time length $T$ equals one, d-GLAD reduces to the GLAD model. Figure \ref{fig:graphDGLAD} depicts the probabilistic graphical model of d-GLAD. The model is demonstrated successful in detecting change in topics of the scientific publications and party affiliation shift of US senators.

%% file: conclusions.tex
In this paper we present a survey of social media anomaly detection methods. Based on the type of target anomalies, these methods fall into two categories: point anomaly detection and group anomaly detection. Moreover, given the different formats of input information, they can be further classified into activity-based approaches and graph-based approaches. For the graph-based approaches, we divide the methodologies according to whether they consider the time dynamics of the social graph. 

One challenge in anomaly detection is to distinguish between data errors and the ``genuine'' anomalies, i.e, those that were caused by the change in the underlying data distribution. As in most cases, it is very difficult to obtain the ground truth labels for the anomalies. We usually ignore the differences before conducting the anomaly detection. Only after we obtain the detection results and perform detailed analysis can we tease out the data error and  recognize the ``genuine'' anomalies.

In summary, social media anomaly detection is still at an early stage. Most existing methods rely heavily on the specific application and self-defined anomalies. Some of the reviewed methods are originally designed for other related purposes, such as community monitoring and proximity tracking, instead of anomaly detection. In addition, many of the existing methods deal with memory-resident graphs, while real life social networks are often too large to fit into the memory. Distributed and online social network anomaly detection are two promising areas. In the case of very large social networks, new techniques for effectively summarizing the entire social network are also needed.

%% file: acknowledge.tex
We would like to thank Dr. Sanjay Chawla from University of Sydney for his encouragement and detailed advice on the formulation of the anomaly categorization, distinguish between the data error and true anomalies. We also dedicate our acknowledgement to  Dr. Huan Liu from Arizona State University who read this survey and provide valuable advice.

%% file: survey.bbl
\begin{thebibliography}{10}

\bibitem{Ahmed:2007vn}
Tarem Ahmed, Tarem Ahmed, Boris Oreshkin, and Mark" Coates.
\newblock Machine learning approaches to network anomaly detection.
\newblock {\em In Proceedings of the second workshop on tackling computer
  systems problems with machine learning(SYSML)}, 2007.

\bibitem{Akoglu2009}
L~Akoglu and M~McGlohon.
\newblock {Anomaly detection in large graphs}.
\newblock {\em In CMU-CS-09-173 Technical}, (November), 2009.

\bibitem{akoglu2012fast}
L.~Akoglu, H.~Tong, J.~Vreeken, and C.~Faloutsos.
\newblock Fast and reliable anomaly detection in categorical data.
\newblock 2012.

\bibitem{akoglu2010oddball}
Leman Akoglu, Mary McGlohon, and Christos Faloutsos.
\newblock Oddball: Spotting anomalies in weighted graphs.
\newblock In {\em Advances in Knowledge Discovery and Data Mining}, pages
  410--421. Springer, 2010.

\bibitem{babbar2013causal}
Sakshi Babbar, Didi Surian, and Sanjay Chawla.
\newblock A causal approach for mining interesting anomalies.
\newblock In {\em Advances in Artificial Intelligence}, pages 226--232.
  Springer, 2013.

\bibitem{bilgin2010dynamic}
C.C. Bilgin and B.~Yener.
\newblock Dynamic network evolution: Models, clustering, anomaly detection.
\newblock Technical report, Technical Report, 2008, Rensselaer University, NY,
  2010.

\bibitem{blei06}
David~M. Blei and John~D. Lafferty.
\newblock Dynamic topic models.
\newblock In {\em Proceedings of the 23rd international conference on Machine
  learning}, ICML '06, pages 113--120, New York, NY, USA, 2006. ACM.

\bibitem{Blei03}
David~M. Blei, Andrew~Y. Ng, and Michael~I. Jordan.
\newblock Latent dirichlet allocation.
\newblock {\em J. Mach. Learn. Res.}, 3:993--1022, March 2003.

\bibitem{autopartPKDD04}
Deepayan Chakrabarti.
\newblock Autopart: parameter-free graph partitioning and outlier detection.
\newblock In {\em Proceedings of the 8th European Conference on Principles and
  Practice of Knowledge Discovery in Databases}, PKDD '04, pages 112--124, New
  York, NY, USA, 2004. Springer-Verlag New York, Inc.

\bibitem{graphLawGenAlgo06}
Deepayan Chakrabarti and Christos Faloutsos.
\newblock Graph mining: Laws, generators, and algorithms.
\newblock {\em ACM Comput. Surv.}, 38(1), June 2006.

\bibitem{Chan_stolfo_1998}
P.K. Chan and S.J. Stolfo.
\newblock Toward scalable learning with non-uniform class and cost
  distributions: A case study in credit card fraud detection.
\newblock In {\em KDD'98}, pages 164--168, 1998.

\bibitem{chandola2007outlier}
V.~Chandola, A.~Banerjee, and V.~Kumar.
\newblock Outlier detection: A survey.
\newblock {\em ACM Computing Surveys, to appear}, 2007.

\bibitem{chandola_TKDD_2010}
Varun Chandola, Arindam Banerjee, and Vipin Kumar.
\newblock Anomaly detection for discrete sequences: A survey.
\newblock {\em IEEE Trans. on Knowl. and Data Eng.}, 24(5):823--839, May 2012.

\bibitem{Chandola:2007kx}
Varun Chandola, Varun Chandola, Arindam Banerjee, and Vipin" Kumar.
\newblock Anomaly detection: A survey.
\newblock 2007.

\bibitem{chawla2006slom}
Sanjay Chawla and Pei Sun.
\newblock Slom: a new measure for local spatial outliers.
\newblock {\em Knowledge and Information Systems}, 9(4):412--429, 2006.

\bibitem{ChengTPK09}
Haibin Cheng, Pang-Ning Tan, Christopher Potter, and Steven Klooster.
\newblock Detection and characterization of anomalies in multivariate time
  series.
\newblock In {\em SDM'09}, pages 413--424, 2009.

\bibitem{SubDue2000}
Diane~J. Cook and Lawrence~B. Holder.
\newblock Graph-based data mining.
\newblock {\em IEEE Intelligent Systems}, 15(2):32--41, March 2000.

\bibitem{das2009detecting}
K.~Das, J.~Schneider, and D.B. Neill.
\newblock {\em Detecting anomalous groups in categorical datasets}.
\newblock Carnegie Mellon University, School of Computer Science, Machine
  Learning Department, 2009.

\bibitem{das2008anomaly}
Kaustav Das, Jeff Schneider, and Daniel~B Neill.
\newblock Anomaly pattern detection in categorical datasets.
\newblock In {\em Proceedings of the 14th ACM SIGKDD international conference
  on Knowledge discovery and data mining}, pages 169--176. ACM, 2008.

\bibitem{Duch_CORR_2004}
Florence Duch{\^e}ne, Catherine Garbay, and Vincent Rialle.
\newblock Mining heterogeneous multivariate time-series for learning meaningful
  patterns: Application to home health telecare.
\newblock {\em CoRR}, abs/cs/0412003, 2004.

\bibitem{Dumouchel99Bayes1-stepMarkov}
William Dumouchel.
\newblock {Computer Intrusion Detection Based on Bayes Factors for Comparing
  Command Transition Probabilities}.
\newblock Technical report, 1999.

\bibitem{anomStuctSimToNormICDM07}
William Eberle and Lawrence Holder.
\newblock Discovering structural anomalies in graph-based data.
\newblock In {\em Proceedings of the Seventh IEEE International Conference on
  Data Mining Workshops}, ICDMW '07, pages 393--398, Washington, DC, USA, 2007.
  IEEE Computer Society.

\bibitem{Erosheva04b}
Elena Erosheva, Stephen Fienberg, and John Lafferty.
\newblock {Mixed-membership models of scientific publications}.
\newblock {\em Proceedings of the National Academy of Sciences of the United
  States of America}, 101(Suppl 1):5220--5227, April 2004.

\bibitem{Eskin02ageometric}
E.~Eskin, A.~Arnold, M.~Prerau, L.~Portnoy, and S.~Stolfo.
\newblock {\em {A geometric framework for unsupervised anomaly detection:
  Detecting intrusions in unlabeled data}}.
\newblock Kluwer, 2002.

\bibitem{findingTribesKDD07}
Lisa Friedland and David Jensen.
\newblock Finding tribes: identifying close-knit individuals from employment
  patterns.
\newblock In {\em Proceedings of the 13th ACM SIGKDD international conference
  on Knowledge discovery and data mining}, KDD '07, pages 290--299, New York,
  NY, USA, 2007. ACM.

\bibitem{Ghosh_schwartzbard_1999}
Anup~K. Ghosh and Aaron Schwartzbard.
\newblock A study in using neural networks for anomaly and misuse detection.
\newblock In {\em Proceedings of the 8th conference on USENIX Security
  Symposium - Volume 8}, SSYM'99, pages 12--12, Berkeley, CA, USA, 1999. USENIX
  Association.

\bibitem{Guralnik_1999}
Valery Guralnik and Jaideep Srivastava.
\newblock Event detection from time series data.
\newblock In {\em Proceedings of the fifth ACM SIGKDD international conference
  on Knowledge discovery and data mining}, KDD '99, pages 33--42, New York, NY,
  USA, 1999. ACM.

\bibitem{Hanneke2010}
Steve Hanneke and Eric~P. Xing.
\newblock Discrete temporal models of social networks.
\newblock In {\em Proceedings of the 2006 conference on Statistical network
  analysis}, ICML'06, pages 115--125, Berlin, Heidelberg, 2007.
  Springer-Verlag.

\bibitem{Hawkins1980}
D.M. Hawkins.
\newblock {\em Identification of outliers}.
\newblock Chapman and Hall, 1980.

\bibitem{Heard:2010zr}
Nicholas~A. Heard, David~J. Weston, Kiriaki Platanioti, and David~J. Hand.
\newblock Bayesian anomaly detection methods for social networks.
\newblock 11 2010.

\bibitem{horn2011online}
C.~Horn and R.~Willett.
\newblock Online anomaly detection with expert system feedback in social
  networks.
\newblock In {\em Acoustics, Speech and Signal Processing (ICASSP), 2011 IEEE
  International Conference on}, pages 1936--1939. IEEE, 2011.

\bibitem{eigenSpaceADKDD04}
Tsuyoshi Id{\'e} and Hisashi Kashima.
\newblock Eigenspace-based anomaly detection in computer systems.
\newblock In {\em Proceedings of the tenth ACM SIGKDD international conference
  on Knowledge discovery and data mining}, pages 440--449. ACM, 2004.

\bibitem{Ihler:2006AED}
Alexander Ihler, Jon Hutchins, and Padhraic Smyth.
\newblock Adaptive event detection with time-varying poisson processes.
\newblock In {\em Proceedings of the 12th ACM SIGKDD international conference
  on Knowledge discovery and data mining}, KDD '06, pages 207--216, New York,
  NY, USA, 2006. ACM.

\bibitem{Vardi99ahybrid}
Wen~H. Ju and Yehuda Vardi.
\newblock A hybrid high-order markov chain model for computer intrusion
  detection.
\newblock {\em Journal of Computational and Graphical Statistics},
  10(2):277--295, 2001.

\bibitem{Keogh2005HotSax}
Eamonn Keogh, Jessica Lin, and Ada Fu.
\newblock Hot sax: Efficiently finding the most unusual time series
  subsequence.
\newblock In {\em Proceedings of the Fifth IEEE International Conference on
  Data Mining}, ICDM '05, pages 226--233, Washington, DC, USA, 2005. IEEE
  Computer Society.

\bibitem{CHRONICLE09}
Min-Soo Kim and Jiawei Han.
\newblock Chronicle: A two-stage density-based clustering algorithm for dynamic
  networks.
\newblock In {\em Proceedings of the 12th International Conference on Discovery
  Science}, DS '09, pages 152--167, Berlin, Heidelberg, 2009. Springer-Verlag.

\bibitem{Kleinberg99}
Jon~M. Kleinberg.
\newblock Authoritative sources in a hyperlinked environment.
\newblock {\em J. ACM}, 46(5):604--632, September 1999.

\bibitem{Kolar10}
Mladen Kolar, Le~Song, Amr Ahmed, and Eric~P. Xing.
\newblock {Estimating time-varying networks}.
\newblock {\em Annals of Applied Statistics}, 4(1):94--123, 2010.

\bibitem{Lakhina2004NetTrafAD}
Anukool Lakhina, Mark Crovella, and Christophe Diot.
\newblock Diagnosing network-wide traffic anomalies.
\newblock {\em SIGCOMM Comput. Commun. Rev.}, 34:219--230, August 2004.

\bibitem{Lappas09}
Theodoros Lappas, Kun Liu, and Evimaria Terzi.
\newblock Finding a team of experts in social networks.
\newblock In {\em Proceedings of the 15th ACM SIGKDD international conference
  on Knowledge discovery and data mining}, KDD '09, pages 467--476, New York,
  NY, USA, 2009. ACM.

\bibitem{Lazarevic:2003ys}
Ar~Lazarevic, Ar~Lazarevic, Aysel Ozgur, Levent Ertoz, Jaideep Srivastava, and
  Vipin" Kumar.
\newblock A comparative study of anomaly detection schemes in network intrusion
  detection.
\newblock {\em In Proceedings of the third SIAM international conference on
  Data Mining}, 2003.

\bibitem{Lin_DMKD_2003}
Jessica Lin, Eamonn Keogh, Stefano Lonardi, and Bill Chiu.
\newblock A symbolic representation of time series, with implications for
  streaming algorithms.
\newblock In {\em Proceedings of the 8th ACM SIGMOD workshop on Research issues
  in data mining and knowledge discovery}, DMKD '03, pages 2--11, New York, NY,
  USA, 2003. ACM.

\bibitem{linkDiscoveryViaRarityICDM03}
Shou-de Lin and Hans Chalupsky.
\newblock Unsupervised link discovery in multi-relational data via rarity
  analysis.
\newblock In {\em Proceedings of the Third IEEE International Conference on
  Data Mining}, ICDM '03, pages 171--, Washington, DC, USA, 2003. IEEE Computer
  Society.

\bibitem{changingSubgraphsICDM08}
Zheng Liu, Jeffrey~Xu Yu, Yiping Ke, Xuemin Lin, and Lei Chen.
\newblock Spotting significant changing subgraphs in evolving graphs.
\newblock In {\em Proceedings of the 2008 Eighth IEEE International Conference
  on Data Mining}, ICDM '08, pages 917--922, Washington, DC, USA, 2008. IEEE
  Computer Society.

\bibitem{Luxburg2007}
Ulrike Luxburg.
\newblock A tutorial on spectral clustering.
\newblock {\em Statistics and Computing}, 17(4):395--416, December 2007.

\bibitem{maruhashi2011multiaspectforensics}
Koji Maruhashi, Fan Guo, and Christos Faloutsos.
\newblock Multiaspectforensics: Pattern mining on large-scale heterogeneous
  networks with tensor analysis.
\newblock In {\em Advances in Social Networks Analysis and Mining (ASONAM),
  2011 International Conference on}, pages 203--210. IEEE, 2011.

\bibitem{weightedGraphPatternKDD08}
Mary McGlohon, Leman Akoglu, and Christos Faloutsos.
\newblock Weighted graphs and disconnected components: patterns and a
  generator.
\newblock In {\em Proceedings of the 14th ACM SIGKDD international conference
  on Knowledge discovery and data mining}, KDD '08, pages 524--532, New York,
  NY, USA, 2008. ACM.

\bibitem{outrank08}
H.~D.~K. Moonesinghe and Pang-Ning Tan.
\newblock Outrank: a {Graph-Based} outlier detection framework using random
  walk.
\newblock {\em International Journal on Artificial Intelligence Tools}, 17(1),
  2008.

\bibitem{muandet2013one}
Krikamol Muandet and Bernhard Sch{\"o}lkopf.
\newblock One-class support measure machines for group anomaly detection.
\newblock {\em stat}, 1050:1, 2013.

\bibitem{noble2003graph}
C.C. Noble and D.J. Cook.
\newblock Graph-based anomaly detection.
\newblock In {\em Proceedings of the ninth ACM SIGKDD international conference
  on Knowledge discovery and data mining}, pages 631--636. ACM, 2003.

\bibitem{Pan:2004:AMC:1014052.1014135}
Jia-Yu Pan, Hyung-Jeong Yang, Christos Faloutsos, and Pinar Duygulu.
\newblock Automatic multimedia cross-modal correlation discovery.
\newblock In {\em Proceedings of the tenth ACM SIGKDD international conference
  on Knowledge discovery and data mining}, KDD '04, pages 653--658, New York,
  NY, USA, 2004. ACM.

\bibitem{park2008scan}
DMY Park, C.E. Priebe, D.J. Marchette, and A.~Youssef.
\newblock Scan statistics on enron hypergraphs.
\newblock {\em Interface}, 2008.

\bibitem{ADinSeriesGraphsARMA05}
Brandon Pincombe.
\newblock Anomaly detection in time series of graphs using arma processes.
\newblock {\em ASOR BULLETIN}, 24(4):2, 2005.

\bibitem{priebe05scan}
Carey~E. Priebe, John~M. Conroy, David~J. Marchette, and Youngser Park.
\newblock Scan statistics on enron graphs.
\newblock {\em Comput. Math. Organ. Theory}, 11(3):229--247, October 2005.

\bibitem{anomalousLinkKDDExplor05}
Matthew~J. Rattigan and David Jensen.
\newblock The case for anomalous link discovery.
\newblock {\em SIGKDD Explor. Newsl.}, 7(2):41--47, December 2005.

\bibitem{Ringberg_2007}
Haakon Ringberg, Augustin Soule, Jennifer Rexford, and Christophe Diot.
\newblock Sensitivity of pca for traffic anomaly detection.
\newblock {\em SIGMETRICS Perform. Eval. Rev.}, 35(1):109--120, June 2007.

\bibitem{yu2014glad}
Yu~Rose, He~Xinran, and Liu. Yan.
\newblock Glad: Group anomaly detection in social media analysis.
\newblock {\em Proceedings of the 20th ACM SIGKDD international conference on
  Knowledge discovery and data mining}, 2014.

\bibitem{Rosen-Zvi04}
Michal Rosen-Zvi, Thomas Griffiths, Mark Steyvers, and Padhraic Smyth.
\newblock The author-topic model for authors and documents.
\newblock In {\em Proceedings of the 20th conference on Uncertainty in
  artificial intelligence}, UAI '04, pages 487--494, Arlington, Virginia,
  United States, 2004. AUAI Press.

\bibitem{schonlau2001computer}
M.~Schonlau, W.~DuMouchel, W.H. Ju, A.F. Karr, M.~Theus, and Y.~Vardi.
\newblock Computer intrusion: Detecting masquerades.
\newblock {\em Statistical Science}, pages 58--74, 2001.

\bibitem{Schonlau2000uniqueness}
Matthias Schonlau and Martin Theus.
\newblock Detecting masquerades in intrusion detection based on unpopular
  commands.
\newblock {\em Inf. Process. Lett.}, 76:33--38, November 2000.

\bibitem{Silva:2008iss}
J.~Silva and R.~Willett.
\newblock Detection of anomalous meetings in a social network.
\newblock In {\em Information Sciences and Systems, 2008. CISS 2008. 42nd
  Annual Conference on}, pages 636 --641, march 2008.

\bibitem{silva2008hypergraph}
J.~Silva and R.~Willett.
\newblock Hypergraph-based anomaly detection in very large networks.
\newblock 2008.

\bibitem{Song05}
Xiaodan Song, Ching-Yung Lin, Belle~L. Tseng, and Ming-Ting Sun.
\newblock Modeling and predicting personal information dissemination behavior.
\newblock In {\em Proceedings of the eleventh ACM SIGKDD international
  conference on Knowledge discovery in data mining}, KDD '05, pages 479--488,
  New York, NY, USA, 2005. ACM.

\bibitem{graphScopeKDD07}
Jimeng Sun, Christos Faloutsos, Spiros Papadimitriou, and Philip~S. Yu.
\newblock Graphscope: parameter-free mining of large time-evolving graphs.
\newblock In {\em Proceedings of the 13th ACM SIGKDD international conference
  on Knowledge discovery and data mining}, KDD '07, pages 687--696, New York,
  NY, USA, 2007. ACM.

\bibitem{neighborFormationADinBGraphICDM05}
Jimeng Sun, Huiming Qu, Deepayan Chakrabarti, and Christos Faloutsos.
\newblock Neighborhood formation and anomaly detection in bipartite graphs.
\newblock In {\em Proceedings of the Fifth IEEE International Conference on
  Data Mining}, ICDM '05, pages 418--425, Washington, DC, USA, 2005. IEEE
  Computer Society.

\bibitem{sun2004local}
Pei Sun and Sanjay Chawla.
\newblock On local spatial outliers.
\newblock In {\em Data Mining, 2004. ICDM'04. Fourth IEEE International
  Conference on}, pages 209--216. IEEE, 2004.

\bibitem{sun2006mining}
Pei Sun, Sanjay Chawla, and Bavani Arunasalam.
\newblock Mining for outliers in sequential databases.
\newblock SIAM, 2006.

\bibitem{Takeuchi_TKDD_2006}
Jun-ichi Takeuchi and Kenji Yamanishi.
\newblock A unifying framework for detecting outliers and change points from
  time series.
\newblock {\em IEEE Trans. on Knowl. and Data Eng.}, 18(4):482--492, April
  2006.

\bibitem{Tong:2006:FRW:1193207.1193363}
Hanghang Tong, Christos Faloutsos, and Jia-Yu Pan.
\newblock Fast random walk with restart and its applications.
\newblock In {\em Proceedings of the Sixth International Conference on Data
  Mining}, ICDM '06, pages 613--622, Washington, DC, USA, 2006. IEEE Computer
  Society.

\bibitem{TongPSYF08}
Hanghang Tong, Spiros Papadimitriou, Jimeng Sun, Philip~S. Yu, and Christos
  Faloutsos.
\newblock Colibri: fast mining of large static and dynamic graphs.
\newblock In {\em Proceedings of the 14th ACM SIGKDD international conference
  on Knowledge discovery and data mining}, KDD '08, pages 686--694, New York,
  NY, USA, 2008. ACM.

\bibitem{proximTrackEvolvBGraphSDM08}
Hanghang Tong, Spiros Papadimitriou, Philip~S. Yu, and Christos Faloutsos.
\newblock {Proximity Tracking on Time-Evolving Bipartite Graphs}.
\newblock 2008.

\bibitem{WSARE3}
Weng-Keen Wong, Andrew Moore, Gregory Cooper, and Michael Wagner.
\newblock Bayesian network anomaly pattern detection for disease outbreaks.
\newblock In Tom Fawcett and Nina Mishra, editors, {\em Proceedings of the
  Twentieth International Conference on Machine Learning}, pages 808--815,
  Menlo Park, California, August 2003. AAAI Press.

\bibitem{xiong2011hierarchical}
L.~Xiong, B.~Poczos, J.~Schneider, A.~Connolly, and J.~VanderPlas.
\newblock Hierarchical probabilistic models for group anomaly detection.
\newblock In {\em Proceedings of International Conference on Artifcial
  Intelligence and Statistics}, 2011.

\bibitem{xiong_group_2011}
Liang Xiong, Barnab{\'a}s P{\'o}czos, and Jeff~G Schneider.
\newblock Group anomaly detection using flexible genre models.
\newblock In {\em NIPS}, pages 1071--1079, 2011.

\bibitem{SCANKDD07}
Xiaowei Xu, Nurcan Yuruk, Zhidan Feng, and Thomas A.~J. Schweiger.
\newblock Scan: a structural clustering algorithm for networks.
\newblock In {\em Proceedings of the 13th ACM SIGKDD international conference
  on Knowledge discovery and data mining}, KDD '07, pages 824--833, New York,
  NY, USA, 2007. ACM.

\bibitem{Yankov2008diskAware}
Dragomir Yankov, Eamonn Keogh, and Umaa Rebbapragada.
\newblock Disk aware discord discovery: finding unusual time series in terabyte
  sized datasets.
\newblock {\em Knowl. Inf. Syst.}, 17:241--262, November 2008.

\bibitem{Yue_Wu_Wang_Li_Chu_2007}
Dianmin Yue, Xiaodan Wu, Yunfeng Wang, Yue Li, and Chao-Hsien Chu.
\newblock A review of data mining-based financial fraud detection research.
\newblock {\em 2007 International Conference on Wireless Communications
  Networking and Mobile Computing}, (C):5514--5517, 2007.

\end{thebibliography}
